\documentclass{article}

\usepackage[final]{neurips_2020}

\usepackage[utf8]{inputenc} 
\usepackage[T1]{fontenc}    
\usepackage{hyperref}       
\usepackage{url}            
\usepackage{booktabs}       
\usepackage{amsfonts}       
\usepackage{nicefrac}       
\usepackage{microtype}      
\usepackage{xcolor}
\usepackage{graphicx}
\usepackage{amsmath}
\usepackage{booktabs}
\usepackage{caption} 
\title{SGoLAM: Simultaneous Goal Localization and Mapping for Multi-Object Goal Navigation}

\author{%
  Junho Kim \qquad Eun Sun Lee \qquad Mingi Lee \qquad Dongsu Zhang \qquad  Young Min Kim \\
Dept. of Electrical and Computer Engineering, Seoul National University, Korea\\
{\tt\small \{ 82magnolia, eunsunlee, mingi1019, 96lives, youngmin.kim \} @snu.ac.kr}

}

\begin{document}

\maketitle

\begin{abstract}
    We present SGoLAM, short for simultaneous goal localization and mapping, which is a simple and efficient algorithm for Multi-Object Goal navigation. 
    Given an agent equipped with an RGB-D camera and a GPS/Compass sensor, our objective is to have the agent navigate to a sequence of target objects in realistic 3D environments. 
    Our pipeline fully leverages the strength of classical approaches for visual navigation, by decomposing the problem into two key components: mapping and goal localization.
    The mapping module converts the depth observations into an occupancy map, and the goal localization module marks the locations of goal objects.
    The agent's policy is determined using the information provided by the two modules: if a current goal is found, plan towards the goal and otherwise, perform exploration.
    As our approach does not require any training of neural networks, it could be used in an off-the-shelf manner, and amenable for fast generalization in new, unseen environments.
    Nonetheless, our approach performs on par with the state-of-the-art learning-based approaches.
    SGoLAM is ranked 2nd in the CVPR 2021 MultiON (Multi-Object Goal Navigation) challenge. We have made our code publicly available at \emph{https://github.com/eunsunlee/SGoLAM}.
\end{abstract}

\section{Introduction}
Visual navigation, the task of navigating a 3D environment using visual information, is a crucial component for autonomous agents.
The development of agents that could find an object in novel environments or move towards a specified goal location via visual navigation can serve as a building block for higher-level services ranging from medical assistant robots to home robots.
Further, the advancements in 3D sensing technology have made a vast range of sensors available for mobile robots, greatly enhancing the range and quality of utilizable information.

In this paper, we tackle the problem of Multi-Object Goal navigation.
Given an agent equipped with an RGB-D camera and a GPS/Compass sensor, the goal of Multi-Object Goal navigation is to plan policies for visiting a designated set of objects in order.
Multi-object goal navigation is an extension of the classical object goal navigation task(\textcolor{blue}{~\cite{Zhu2016})}), where multiple instead of single object goals should be found. 
Such additional difficulty aims to rigorously evaluate two main components in visual navigation: goal localization and mapping. 
The task assesses the agent's ability to locate the target objects and generate appropriate maps that can cache useful spatial information.

We present SGoLAM, a simple and efficient algorithm for Multi-Object Goal navigation. 
The key idea is to localize goals and simultaneously map the environment using classic projective geometry in a modular fashion.
Each module in SGoLAM addresses a specific task such as goal localization, mapping, and planning.

The benefit of SGoLAM is that the approach can be deployed on an agent without any training procedure, in contrast to learning-based methods.
Many end-to-end learning-based approaches which directly learn the navigation policy from sensory data lack the generalizability over various environments and transferability over tasks. 
SGoLAM presents work robust to previously unseen RGB-D inputs and recyclable modules for other navigation tasks.

We validate SGoLAM for Multi-Object Goal navigation in Habitat simulator (\textcolor{blue}{~\cite{habitat2020sim2real}}) with 3D indoor scene datasets from Matterport3D (\textcolor{blue}{~\cite{Matterport3D}}). 
Our approach significantly outperforms the baseline methods presented in the CVPR 2021 MultiON challenge and performs on par with the state-of-the-art learning-based approaches.

\section{Related Work}
\subsection{Embodied AI \& Visual Navigation}
Enabled by the availability of realistic 3D environments and simulation platforms for robotic agents, significant progress has been made in the field of embodied AI in the past few years.
Creative work on visual navigation has been done with some common themes like egocentric perception, long-term planning, learning from interaction, and holding a semantic understanding of an environment.
As an output of this work, a plethora of task definitions have been made and then converged to several common definitions regarding the nature of the tasks (\textcolor{blue}{\cite{Anderson2018}}).

Navigation tasks can be distinguished in many dimensions but the nature of a task depends mostly on the type of goal.
In PointGoal navigation, the agent must navigate to a specific location given relative to where the agent is currently positioned (\emph{e.g.}, (150,300)).
PointGoal task in simulated environments with noiseless sensors has recently been solved with near-perfect performance (\textcolor{blue}{\cite{wijmans2020ddppo}}) and shifted focus to real world deployment with noisy sensors (\textcolor{blue}{\cite{habitat2020sim2real}}, \textcolor{blue}{\cite{ramakrishnan2020occupancy}}, \textcolor{blue}{\cite{chaplot2020learning}}).
In ObjectGoal navigation, the agent must navigate to an object of a specific category drawn from a predefined set (\emph{e.g.}, 'chair').
Despite being formulated in early work (\textcolor{blue}{\cite{Zhu2016}}), the task remains far from being solved.
There have been various approaches tackling ObjectGoal navigation by building episodic semantic map (\textcolor{blue}{\cite{Chaplot2020}}), exploiting object relationships (\textcolor{blue}{\cite{Qiu2020}}), and by learning spatial context (\textcolor{blue}{\cite{Druon2020}}).
In AreaGoal navigation, the agent must navigate to a room of a specified category (\emph{e.g.}, 'kitchen').
AreaGoal has been formulated first in (\textcolor{blue}{\cite{mattersim}}) as visually-grounded natural language navigation task and has been dealt with formulating agent under Bayesian filtering (\textcolor{blue}{\cite{anderson2019chasing}}) and contextual global graphical planner (\textcolor{blue}{\cite{Deng2020}}).

As the performance of proposed methods in each task is increasing incredibly fast, recent discussion in the area is starting to tackle complex long-horizon tasks.
Room-scale environments are expanded to building-scale environments and single goal tasks are expanded to sequential goals.
Considering the practical use of an indoor embodied robot, Multi-Object Goal navigation is one of the most important tasks in embodied AI.
\textcolor{blue}{\cite{fang2019scene}} proposes 'object-search' where agent must navigate to some categories of object with no specific order.
In a static scene, the position and the number of goals would be fixed which limits task complexity and property.
\textcolor{blue}{\cite{beeching2019deep}} proposes an 'ordered k-item' task where the agent must navigate to a set of items in a specified order fixed across episodes.
\textcolor{blue}{\cite{Wani2020}} proposes MultiON task where an agent must navigate to arbitrary colored objects given as an episode-specific ordered set.
Episode-specificity requires grounding object class labels to their visual appearance.
The ordered aspect requires an efficient strategy to store information of possible future goals.
To this extent, we choose to tackle Multi-Object Goal navigation following the definition of the MultiON task.

\subsection{Mapping in Navigation}
Long-horizon navigation task in a complex realistic environment is considered beyond the capability of existing memory-less systems.
Including revisits to SLAM methods, there is a growing interest in extending the agents with memory structures.
A simple implicit type of memory has been studied in reinforcement learning settings with memory-base policy using RNN (\textcolor{blue}{\cite{oh2016control}}, \textcolor{blue}{\cite{mirowski2017learning}}, \textcolor{blue}{\cite{mousavian2019visual}}).
Drawbacks of such policy are apparent as merging observations into lower dimension state vectors can easily lose information and optimizing over long sequences is difficult in backpropagation through time.
External memory structure can be categorized broadly into topological maps and spatial maps.
Topological map(\textcolor{blue}{\cite{savinov2018semiparametric}}, \textcolor{blue}{\cite{mirowski2017learning}}) is a more generic type of memory which stores landmarks(\emph{e.g.}, specific input frame) as nodes and their connectivity as edges.
Spatial maps are mostly in the form of 2D grids where dimensions align with an environment's top-down layout.
Starting with SLAM which builds the very basic form of spatial maps, there have been extensive studies on the use of spatial map(\textcolor{blue}{\cite{gordon2018iqa}}, \textcolor{blue}{\cite{8578982}}, \textcolor{blue}{\cite{zhang2020neural}}, \textcolor{blue}{\cite{chaplot2020learning}}, \textcolor{blue}{\cite{Chaplot2020}}) and has recently shown the state-of-the-art performance on the ObjectGoal navigation tasks(\textcolor{blue}{\cite{Chaplot2020}}).

\textcolor{blue}{\cite{Wani2020}} provides baseline agents for Multi-Object Goal navigation, each adopted from representative methods utilizing either implicit or external memories.
However, all baselines are based on learning-based methods and the best performing method adopted from \textcolor{blue}{\cite{8578982}} requires a heavy, complex neural image feature map.
In contrast, our method does not need any training and has a simple, straightforward structure while outperforming all of the baselines.

\begin{figure}
\begin{center}
   \includegraphics[width=\linewidth]{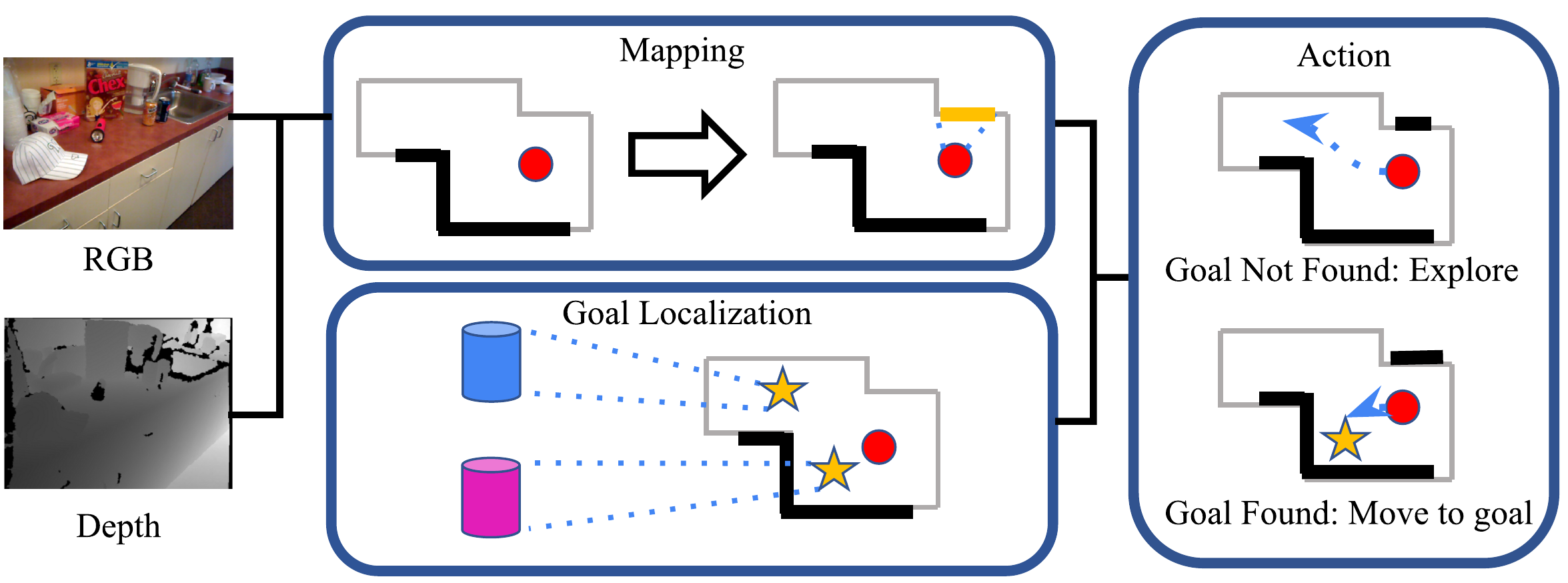}
\end{center}
   \caption{Overview of SGoLAM (simultaneous goal localization and mapping). In the mapping module, RGBD information is first processed to form an occupancy map. In the goal localization module, goals are detected and further projected to form a goal map. The information from these to modules is fused to determine a policy: plan toward goal if goal is found and explore otherwise.}
\label{method}
\end{figure}

\section{Method}
We propose a modular navigation approach, SGoLAM, composed of three components: Mapping module, Goal Localization module, and a Policy module. 
The overview of our approach is visualized in Figure~\ref{method}.
Given the agent's pose from a noiseless GPS and Compass sensor, the mapping module creates the map of the environment by back-projecting the 3D coordinates from the current depth observation. The Goal Localization module detects the target objects from the current RGB observation and creates a goal map by back-projecting the aligned goal location from the depth observation. The Policy module takes in the predicted map and the agent pose from the Mapping module and goal locations from the Goal Localization module when the targets are detected. Based on the localization result, the Policy module outputs actions to explore the unseen area or to navigate to the goal locations. 

\subsection{Task Description}
MultiON task extends the ObjectGoal navigation task, where the objective is to find multiple target objects in a given sequential order. We follow the setup from CVPR 2021 MultiON challenge(\textcolor{blue}{\cite{Wani2020}}) . A set of three target objects are randomly sampled without replacement from 8 cylinders with identical shapes but with different colors: red, green, blue, cyan, magenta, yellow, black, white. The embodied agent is shaped with a cylindrical body of height 1.5m and radius 0.1m and is equipped with an RGB-D camera and a noiseless GPS/Compass sensor. Each episode begins with the agent randomly positioned in an unseen environment. The agent takes one of three navigational actions (\texttt{Move Forward}, \texttt{Turn Right}, and \texttt{Turn Left}) and indicates goal discovery by calling a \texttt{Found} action within 1.5m from its target object. The episode terminates with success when all goals are properly discovered or with failure if the agent incorrectly calls the \texttt{Found} action. 

\subsection{Mapping}
The Mapping module maintains the agent's position $(x, y, \theta)$ and a global occupancy map $O \in \mathbb{R}^{M\times M\times 1}$.  Given a depth image $D$, the module projects the 3D coordinates to an egocentric top-down grid map $m_{t} \in \mathbb{R}^{N\times N\times 1}$ which indicates the probability of the corresponding location being vacant, occupied or unexplored at each time step. Note that $M$ and $N$ denote the size of the global and egocentric map. Based on the current agent's pose and orientation $(x, y, \theta)$ , the egocentric map $m_{t}$ is transformed to an allocentric top-down grid map $a_{t} \in \mathbb{R}^{M\times M\times 1}$. The allocentric top-down map is overlaid on the global occupancy map $O_{t-1}$ from the previous time step to generate a new global occupancy map $O_{t}$. The global map is further used in the Policy module.

\subsection{Goal Localization}
Given an RGB image $I$ and depth map $D$, the Goal Localization module caches the location of target objects.
To prevent redundant exploration, all target objects are localized regardless of whether or not it is the current goal.
The module initiates by detecting target objects, namely colored cylinders, from the image $I$.
Regions in the image that share a similar color with the target objects are marked and further projected to form a goal map, which is a map that stores the 2D location of localized goals.
Specifically, suppose that from $n$ cylinders with color $C=\{c_1, \dots, c_n\}$, the agent must find $k$ cylinders with color $S=\{c_{i_1}, \dots, c_{i_k}\}$.
For each $c_i \in C$, pixel locations $(x, y)$ in the image that satisfy $\|I(x, y) - c_i\| < \epsilon$ for a small constant $\epsilon$ are first marked as putative target objects.
As such naive thresholding is prone to false positives, we further apply connected components labeling(\textcolor{blue}{\cite{connect_components}}) and remove regions whose component size is below a threshold $\delta$.
The filtered pixel locations are stored in a $M \times M$ goal map $G \in \mathbb{R}^{M\times M\times n}$, instantiated as a top-down grid map, where the location of each pixel within the goal map is determined using the depth image $D$.
Note that $n$ grid maps are stored in $G$, with each map storing the goal location for a specific target object.
The goal map $G$ is further used to selecting the appropriate policy for Multi-Object Goal navigation.

\subsection{Policy}
The agent policy is determined in the Policy module using the occupancy map $O$ and goal map $G$.
If the current object goal is localized within $G$, the agent plans toward the goal, and otherwise the agent performs exploration.
To elaborate, suppose the current object goal is the cylinder with color $c_i$.
The agent first inspects the goal map corresponding to the i\textsuperscript{th} object goal.
If there are non-zero pixels within the map, the agent performs actions to move closer to the goal location, which is estimated as the mean pixel location of all non-zero pixels.
The specific actions are planned with the D* algorithm(\textcolor{blue}{\cite{dstar}}), which is a variant of the A* algorithm that is suitable for dynamic environments where the map constantly changes.
If the map is left empty, the agent explores the environment with the frontier-based method(\textcolor{blue}{\cite{frontier}}) that leverages information from $O$ until the current object goal is found.
The frontier-based method is a classical method for robot navigation, where the agent is encouraged to move towards regions in the occupancy map with the largest intersections with the unexplored area.
Note that both the mapping and goal localization modules operate simultaneously with the execution of agent policies.
This allows SGoLAM to rapidly detect target object goals and minimize redundant exploration.

\section{Experimental Results}
\subsection{Experimental Setup}
SGoLAM is mainly implemented using PyTorch(\textcolor{blue}{\cite{pytorch}}), and is accelerated with a single RTX 2080 GPU.
All evaluations are performed on the Habitat simulator(\textcolor{blue}{\cite{savva2019habitat}}) using the Matterport3D(\textcolor{blue}{\cite{Matterport3D}}) scenes.
We set both the goal map and occupancy map size to be $M,N=550$.
Also, the threshold values used for goal localization is as follows: $\epsilon=0.001, \delta=50$.
We compare SGoLAM against four baselines:

\texttt{NoMap(RNN)}: An agent that does not utilize any map information.
An RNN encoder keeps track of the agent's history, and the agent makes actions using the hidden state.

\texttt{ProjNeuralMap}: This agent projects image features onto a top-down grid map, and utilizes this map information for Multi-Object Goal navigation.
The image features are obtained by passing RGBD observations through a pre-trained CNN.

\texttt{AuxTaskMap}: This agent shares the same neural network architecture as the \texttt{ProjNeuralMap} agent.
However, it further fine-tunes the feature extractor CNN with three auxiliary tasks: goal location estimation, goal visibility estimation, and goal distance estimation.
The agent is trained on these tasks in a supervised manner.  

\texttt{VisMemoryMap}: This agent utilizes an array of memory vectors to keep track of salient past observations, similar to Memory Networks (\textcolor{blue}{\cite{memory_net}}).
The memory vectors are used to plan trajectories for Multi-Object Goal navigation.

\subsection{Metrics}
We evaluates the performance of our agent with metrics suggested for object goal navigation task and extended for Multi-Object Goal navigation task in previous work (\textcolor{blue}{\cite{Anderson2018}},\textcolor{blue}{\cite{Wani2020}}).

\textbf{Success}: Binary indicator of success for each episode. The metric is a success if an agent calls \{Found\} for all target objects within a threshold distance, in a correct sequence, and within the allowed maximum steps for each episode. The episode fails if the agent reaches its maximum step without finding all three objects or incorrectly calls \texttt{Found}. 

\textbf{Progress}: The proportion of object goals discovered successfully. In one object goal navigation task, progress evaluates the same metric as success. 

\textbf{SPL}: Extended version of ‘Success weighted by Path Length’. (\textcolor{blue}{\cite{Anderson2018}}). 
\begin{align}
    \text{SPL} = s\cdot d/ max(p,d) 
\end{align}
$s$ is the binary success indicator, $p$ is the total number of steps progressed by the agent and $d = \sum_{i=1}^{n} d_{i-1, i} $, the total geodesic distance from the starting position through each goal location. 

\textbf{PPL}: Progress weighted by Path Length. 
\begin{align}
    \text{PPL} = \bar{s}\cdot \bar{d}/ max(p,\bar{d}) 
\end{align}
$\bar{s}$ is a progress.  $d = \sum_{i=1}^{l} d_{i-1, i} $, where l is the number of objects found. $p$ and $d_{i-1, i} $ are equally defined as before. The metric prevents unfair high weights on shorter trajectory between goals by weighting overall distance based on progress. PPL for one object goal navigation is equal to SPL.

\subsection{Performance Analysis}
\begin{table}[]
\centering
\begin{tabular}{@{}l|c|c|c|c@{}}
\toprule
Method & Success & Progress & PPL & SPL \\ \midrule
NoMap (RNN) & 0.05 & 0.19 & 0.13 & 0.03 \\
ProjNeuralMap & 0.12 & 0.29 & 0.16 & 0.06 \\
VisMemoryMap & 0.43 & 0.57 & 0.36 & 0.27 \\
AuxTaskMap & 0.57 & 0.70 & \textbf{0.45} & \textbf{0.36} \\
SGoLAM & \textbf{0.62} & \textbf{0.71} & 0.39 & 0.34 \\ \bottomrule
\end{tabular}
\caption{Quantitave comparisons with the baseline methods. Note that all metrics are reported in decimals. SGoLAM performs on par with the state-of-the-art learning-based methods. In terms of the overall success rate, SGoLAM outperforms all the baselines by a large margin.}

\label{quant}
\end{table}

Quantitative comparisons with the baselines are shown in Table~\ref{quant}.
SGoLAM performs competitively against the learning-based approaches, and SGoLAM outperforms all other methods in metrics that evaluate the overall success rate, namely success and progress.
However, for metrics that also consider path efficiency (PPL and SPL), the state-of-the-art learning-based method \texttt{AuxTaskMap} outperforms SGoLAM.
This could be attributed to the intrinsic limitations of classical planning-based navigation methods.
As noted by \textcolor{blue}{\cite{classical}}, classical methods for visual navigation tend to show higher success rates than their learning-based counterparts but produce more inefficient paths.
Although such conclusions were made for point-goal navigation (\textcolor{blue}{\cite{wijmans2020ddppo}}), a similar conclusion could be made for Multi-Object Goal navigation.
Nonetheless, the performance of SGoLAM is not far behind that of learning-based approaches without the help of training, making it amenable for fast, effective adaptation in novel environments. 

\section{Conclusion}
In this paper, we introduced SGoLAM, a simple and modular model using classic projective geometry for Multi-Object Goal navigation. The proposed method leverages the benefits of classical navigational approaches which is generally strong and easily adaptable to new, unseen environments without any training requirements. However, this is not to say that the classical approach is the only promising direction of research over learning-based methods in navigation tasks. We believe that there is plenty of room for improvement by taking the complementary advantages of classical and learning-based methods.

\footnotesize{
\bibliographystyle{abbrvnat}
\bibliography{egbib}
}
\end{document}